\documentclass[10pt,twocolumn,letterpaper]{article}

\usepackage[pagenumbers]{wacv} 

\usepackage{graphicx}
\usepackage{booktabs}

\usepackage{amssymb}
\usepackage{amsmath}
\usepackage{amsmath,amssymb,amsfonts}
\usepackage{algorithmic}
\usepackage{graphicx}
\usepackage{textcomp}
\usepackage{booktabs}
\usepackage{multirow}
\usepackage{xcolor}

\definecolor{wacvblue}{rgb}{0.21,0.49,0.74}
\usepackage[pagebackref,breaklinks,colorlinks,allcolors=wacvblue]{hyperref}

\def\wacvPaperID{1853} 
\def\confName{WACV}
\def\confYear{2027}

\title{ACorD: Adaptive Correspondence Discovery for Medical Image Registration}

\author{Arghavan Rezvani\\
University of California, Irvine\\
{\tt\small rezvanid@uci.edu}
\and
Kun Han\\
University of California, Irvine\\
{\tt\small khan7@uci.edu}
\and 
Xiaohui Xie\\
University of California, Irvine\\
{\tt\small xhx@uci.edu}
}

\begin{document}
\maketitle
\begin{abstract}
Medical image registration is a fundamental task for aligning anatomical structures across images. 
Exhaustively searching for point-wise correspondences is prohibitively expensive in both computation and memory, particularly for high-resolution 3D medical images.
Most recent learning-based methods directly predict correspondences in a coarse-to-fine manner, but often propagate intermediate estimates across scales without explicitly weighting correspondence evidence, which can allow coarse-scale errors to persist during refinement.
To address these challenges, we propose ACorD, an adaptive correspondence discovery framework designed to mitigate error propagation with an attention-guided process.
ACorD first predicts multiple potential correspondence candidates through Correlation-guided Cross-scale Offset Prediction (CCOP), which combines correlation maps as spatial matching priors with cross-scale features as global deformation context.
Then, Multi-Head Deformable Aggregation (MHDA) 
aggregates these candidate correspondences using attention, producing a learned, non-uniform weighting over competing matches rather than treating them equally.
The multi-head design further captures complementary deformation patterns, ensuring both global alignment and local precision.
By integrating adaptive offset prediction with selective, attention-based aggregation of correspondence evidence, ACorD enables accurate deformation refinement.
Extensive experiments on four public medical registration benchmarks demonstrate that adaptive correspondence discovery within a multi-scale refinement framework improves registration robustness across diverse anatomical structures.

\end{abstract}

\section{Introduction}
\label{sec:intro}
\label{sec:intro}
Medical image registration \cite{maintz1998survey,sotiras2013deformable,ferrante2017slice}
aligns images across time points, modalities, or subjects by establishing anatomical correspondences. It supports applications such as disease monitoring, treatment planning, image-guided surgery, and population studies, but remains challenging across both large deformations and subtle local misalignments.

\begin{figure}
\centering
\includegraphics[width=0.6\linewidth]{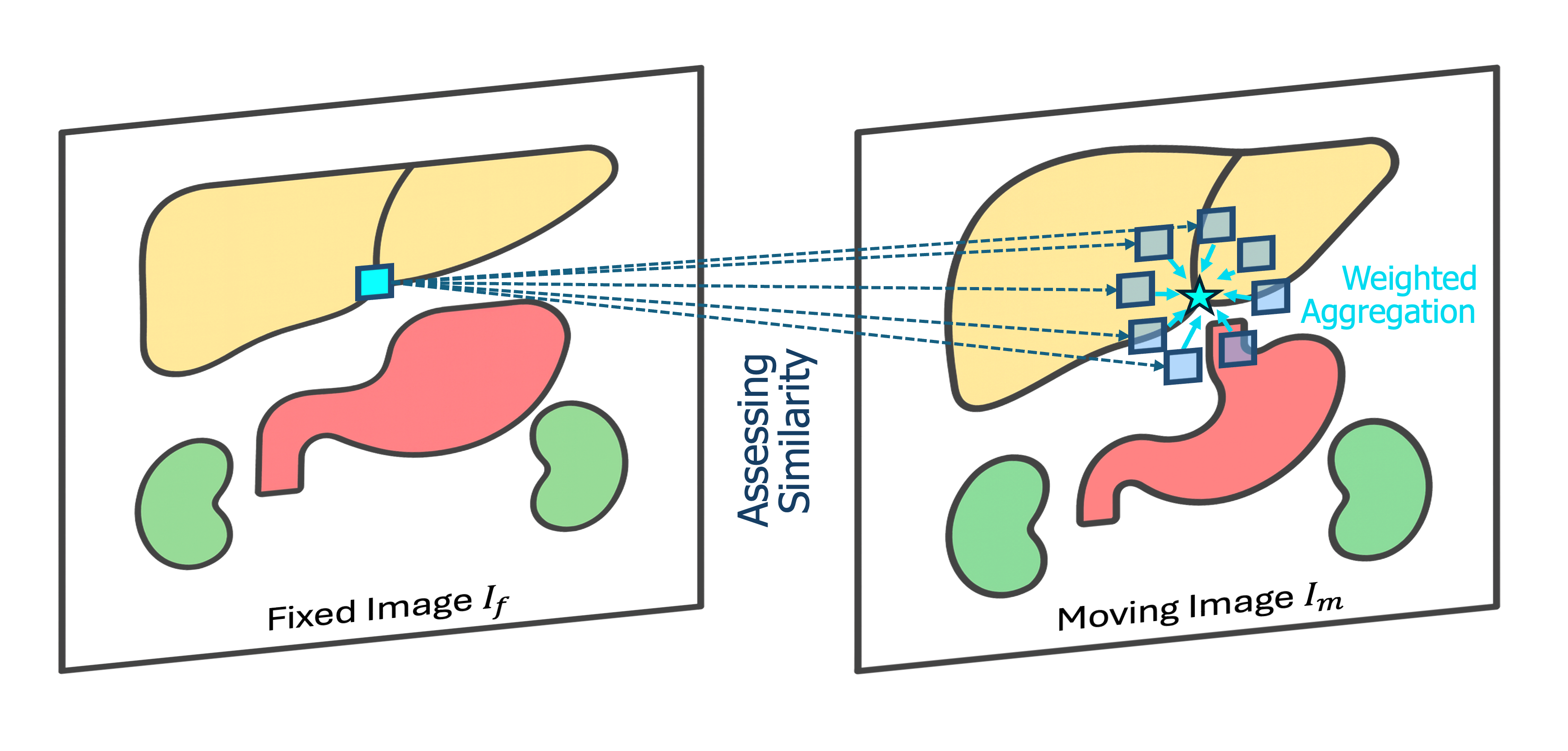}

\caption{\textbf{Adaptive Correspondence Discovery.} ACorD calculates attention between a feature from the fixed image and a set of sampled features from the moving image using predicted correspondence candidates. Predicted offsets guide the sampling, and the resulting attention weights determine the final correspondence.}
\label{fig:three_methods}
\end{figure}

Traditional registration methods optimize image similarity with smoothness constraints \cite{bajcsy1989multiresolution,rueckert2002nonrigid,shen2002hammer,beg2005computing,avants2008symmetric,modat2010fast}, but are computationally costly, initialization-sensitive, and challenged by complex deformations. 
Deep learning methods \cite{fu2020deep,chen2025survey} instead learn registration mappings end-to-end, improving efficiency and performance across applications.
Deep learning registration began with VoxelMorph \cite{balakrishnan2019voxelmorph}, establishing encoder-decoder architectures \cite{dalca2019unsupervised,hoopes2021hypermorph,chen2021vit,chen2022transmorph,shi2022xmorpher,chen2023transmatch}, with multi-scale coarse-to-fine frameworks \cite{mok2020large,meng2022non,kang2022dual,ma2023pivit,meng2023non,zhou2023self,wang2024recursive,meng2024advancing,wang2024pyramid} emerging as the dominant paradigm. 
However, they often propagate deformation estimates across scales without explicitly using correspondence evidence, allowing coarse-level errors to persist during refinement.
Recent methods such as Dual-PRNet++~\cite{kang2022dual}, Im2Grid~\cite{liu2022coordinate}, ModeT~\cite{wang2023modet}, CorrMLP\cite{meng2024correlation}, and SACB-Net~\cite{cheng2025sacb} try to address this limitation by incorporating correlation layers to model voxelwise correspondences, but remain limited to fixed local neighborhoods, typically $3 \times 3 \times 3$, limiting their ability to capture distant correspondences for \textit{large deformations} or adapt search granularity for precise \textit{sub-voxel refinements}.

The core challenge lies in balancing computational efficiency with correspondence flexibility. Fixed neighborhoods are efficient but rigid, while larger searches increase cost and ambiguity. Existing methods lack an adaptive strategy to selectively allocate resources to likely correspondence regions while dynamically adjusting search granularity to the deformation scale.

We propose a novel framework that addresses these limitations through attention-guided deformation propagation with adaptive correspondence discovery (ACorD). The key idea is that robust multi-scale registration requires deformation propagation guided by correspondence evidence rather than blind refinement. ACorD predicts multiple candidate correspondences and uses sparse attention to weight them. Learned offsets shift candidates beyond a fixed local neighborhood according to the observed deformation, avoiding exhaustive 3D correspondence search.

The main contributions of this work are:

\textbf{(1) Attention-guided propagation paradigm}: 
We introduce the first framework that employs 
attention mechanism to weight candidate correspondences before cross-scale propagation, transforming coarse-to-fine registration from unweighted field estimation to learned, similarity-driven aggregation.
\textbf{(2) Adaptive correspondence discovery beyond fixed local windows}: ACorD decouples candidate generation from aggregation: (1) Correlation-guided Cross-scale Offset Prediction (CCOP) leverages spatial correlation priors and cross-scale deformation context to predict multiple candidate locations, 
with learned offsets adaptively shifting the sampled candidate set relative to a reference neighborhood
and (2) Multi-Head Deformable Aggregation (MHDA) then applies sparse attention at these candidate locations, using multi-head aggregation 
to weight candidates non-uniformly across the full deformation spectrum.
 \textbf{(3) State-of-the-art performance across deformation scales}: Extensive experiments on four medical datasets demonstrate consistent improvements-from datasets requiring precise local alignment to the challenging AbdomenCT dataset with large deformations—validating that adaptive and attention-guided registration excels across the entire deformation spectrum.

\section{Background and Related Works}
Let $\Omega\subset\mathbb{R}^3$ denote the spatial domain and $I_m,I_f:\Omega\rightarrow\mathbb{R}$ the moving (source) and fixed (target) images, sampled on an $H\times W\times D$ voxel grid. Registration estimates a transformation $\phi:\Omega\rightarrow\Omega$ mapping fixed-image coordinates to corresponding moving-image locations, such that $(I_m\circ\phi)(\mathbf{x})=I_m(\phi(\mathbf{x}))$. We optimize
    $\phi^{*} = \arg \min_{\phi} \; \mathcal{L}_{\text{sim}}(I_f, I_m \circ \phi) + \lambda \, \mathcal{L}_{\text{reg}}(\phi)$
    ,
   where $\mathcal{L}_{\mathrm{sim}}$ measures image similarity and $\mathcal{L}_{\mathrm{reg}}$ regularizes the transformation.
\label{related_works}

\textbf{Deep Learning-based Image Registration: }
Deep learning enables fast, end-to-end deformable registration. VoxelMorph \cite{balakrishnan2019voxelmorph} established the unsupervised paradigm, later extended with cycle consistency \cite{kim2021cyclemorph} and probabilistic modeling \cite{dalca2019unsupervised}. Transformer methods \cite{chen2022transmorph,zhu2022swin,shi2022xmorpher,ghahremani2024h} further model long-range dependencies, but mostly regress single-scale deformation fields, limiting simultaneous large-displacement and fine-detail alignment.

\textbf{Multi-Scale Coarse-to-Fine Registration:}
Coarse-to-fine methods progressively refine deformation fields across resolutions to handle large displacements and local refinements.
Early deep models like LapIRN \cite{mok2020large} adopted pyramid structures for iterative deformation refinement, followed by more efficient approaches such as NICE-Net \cite{meng2022non}, SD-Net \cite{zhou2023self}, and PC-Reg \cite{yin2023pc}. Recursive pyramid designs in RDPNet \cite{wang2024recursive} and IIRP-Net \cite{ma2024iirp} further support dense multi-scale feature fusion.
More recently, transformer-based methods such as PIViT \cite{ma2023pivit} and NICE-Trans \cite{meng2023non} integrated hierarchical attention to model long-range dependencies. 
These methods typically propagate intermediate estimates across scales, so coarse errors can affect later refinement. ACorD instead forms each refinement from a learned weighting of candidate correspondence evidence before propagation to the next scale.

\textbf{Local Correlation Based Approaches:}
Voxel-wise correlation improves registration, particularly for large deformations, as popularized by RAFT \cite{teed2020raft}. Due to the memory demands of 3D medical data, recent methods adopt local window-based correlation. Dual-PRNet++ \cite{kang2022dual} integrated multi-scale features and correlation maps. im2grid \cite{liu2022coordinate} and ModeT \cite{wang2023modet} captured correspondences within local windows, with ModeT using transformer-based attention. CorrMLP \cite{meng2024correlation} improved deformation learning by employing multi-scale MLP blocks for correlation modeling, while SACB-Net \cite{cheng2025sacb} used spatial-awareness convolutions to emphasize local structural context during matching. Although these methods enhance local matching, they are inherently constrained by fixed-size windows, which limits their ability to capture correspondences over long distances. ACorD instead learns offsets that shift candidate locations beyond the reference neighborhood, enabling adaptive search.

\textbf{Deformable Modeling:}
Deformable modeling enables networks to adaptively sample spatial locations based on content, a concept first introduced in DCN \cite{dai2017deformable}. 
This principle has been extended to attention mechanisms, such as in Deformable DETR \cite{zhu2020deformable} and Deformable ViT \cite{xia2022vision}, which apply deformable sampling to attention queries. 
In medical image registration, TransDCA \cite{chen2023deformable} adopts deformable cross-attention to enhance feature extraction. However, these approaches primarily use deformable sampling to construct stronger feature representations, after which a separate prediction head estimates the deformation field. In contrast, ACorD uses deformable sampling directly for correspondence discovery: CCOP predicts multiple candidate correspondence locations, and MHDA weights them to refine the deformation field.
This design lets the attention weights modulate the contribution of competing correspondences,
enabling adaptive refinement across both large displacements and fine local misalignments.

\section{Methodology}

\subsection{Architecture Overview}

\begin{figure*}[ht]
\centering
\includegraphics[width=\linewidth]{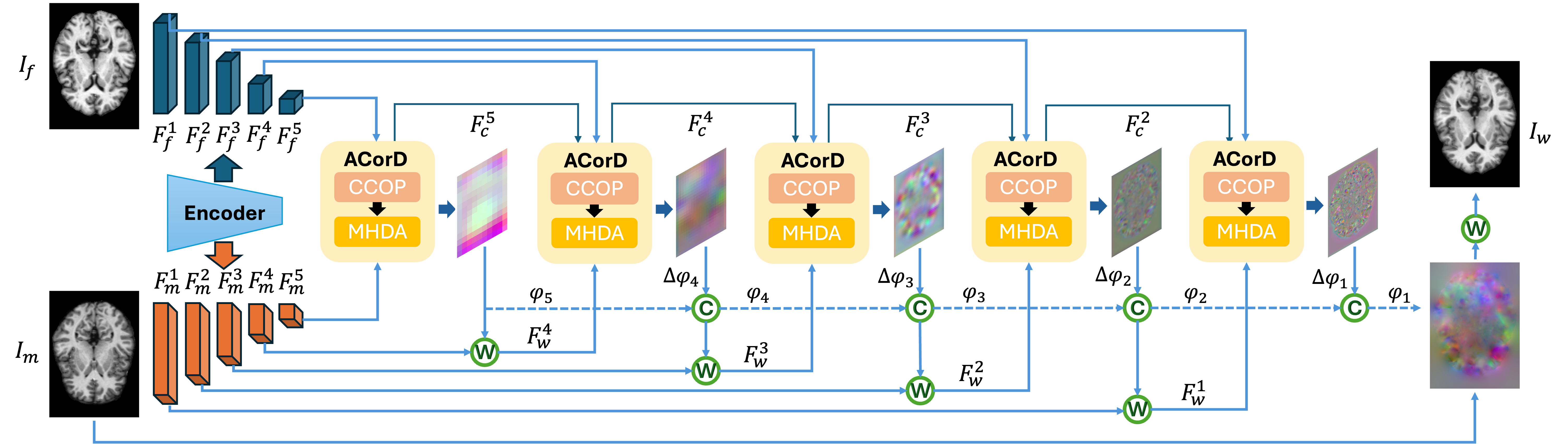}
\caption{\textbf{Overview of the Proposed Method.} A shared encoder extracts multi-scale representations from both images, which are processed by ACorD blocks in a coarse-to-fine manner. At each scale, ACorD combines Correlation-guided Cross-scale Offset Prediction (CCOP) with Multi-Head Deformable Aggregation (MHDA) to refine the deformation field through attention-guided correspondence discovery and propagation.}
\label{fig:fig1}
\end{figure*}


Fig.~\ref{fig:fig1} illustrates the proposed \textbf{Adaptive Correspondence Discovery (ACorD)} framework, comprising:
(1) a shared feature pyramid encoder that extracts hierarchical multi-scale representations from moving and fixed images;  
(2) ACorD blocks that perform adaptive correspondence discovery and selective aggregation at each scale; and
(3) coarse-to-fine deformation refinement across scales. 

Rather than directly propagating each intermediate deformation field, ACorD refines it using learned correspondence weighting before passing it to the next finer scale, reducing the influence of poorly matched candidates during refinement.

\begin{figure*}[t]
\centering
\includegraphics[width=\linewidth]{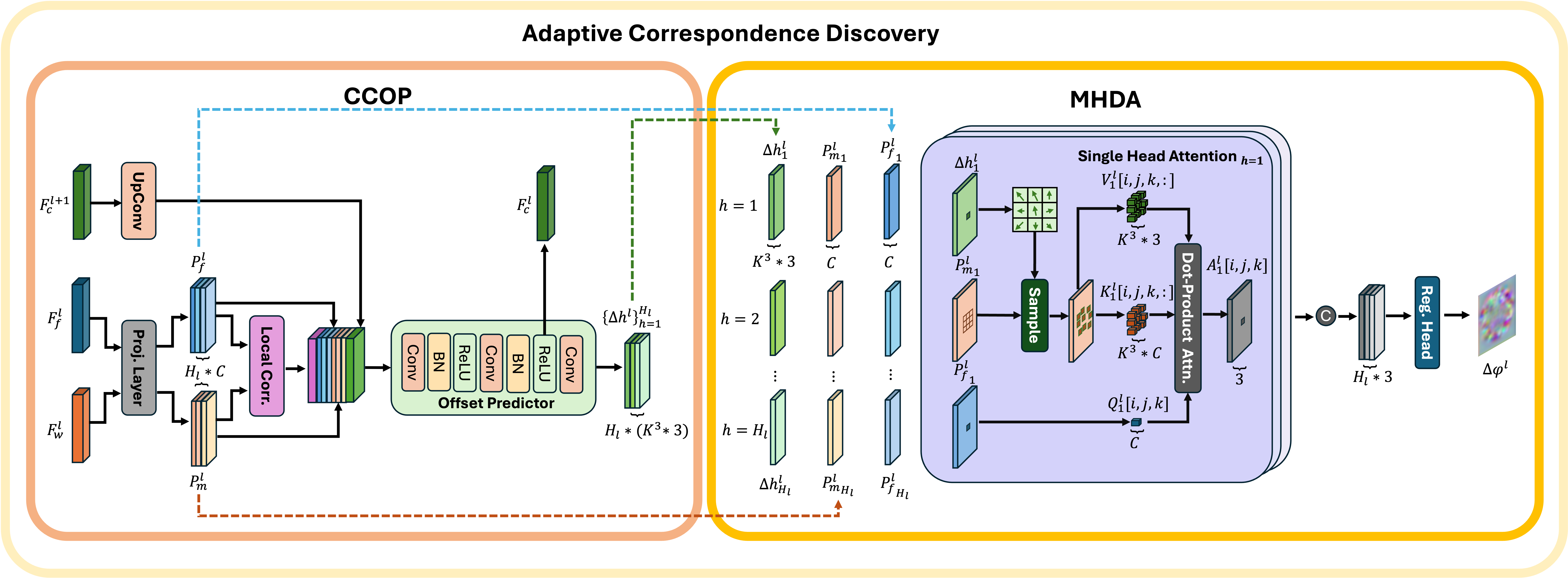}
\caption{
\textbf{Adaptive Correspondence Discovery (ACorD).} 
ACorD consists of two modules: (1) \textbf{CCOP} predicts multi-head adaptive spatial offsets from projected features, local-window correlations, and cross-scale context, while generating features for the next scale.
(2) \textbf{MHDA} uses these offsets to dynamically place attention keys, where queries come from fixed features, keys from warped moving features, and values from sampled positions. 
Attention weights modulate the contribution of each candidate 
and the multi-head design captures diverse deformation patterns.
}
\label{fig:fig2}
\end{figure*}

\subsection{Multi-Scale Feature Extraction}

The shared encoder $\mathcal{E}$ extracts a hierarchy of $L$ feature maps from the moving and fixed images:
\begin{equation}
\{F_m^{(l)}\}_{l=1}^{L} = E(I_m), \qquad
\{F_f^{(l)}\}_{l=1}^{L} = E(I_f).
\end{equation}
where $F_m^{(l)}, F_f^{(l)} \in \mathbb{R}^{C_l \times H_l \times W_l \times D_l}$ denote features at scale $l$. 
Each scale has spatial resolution $\left[\frac{H}{2^{l-1}}, \frac{W}{2^{l-1}}, \frac{D}{2^{l-1}}\right]$ and channel dimension $C_l = C_1 \times 2^{l-1}$, so that deeper levels capture increasingly rich semantic representations.  
Each encoder stage uses two convolutional layers with instance normalization and LeakyReLU activation.  
The resulting multi-scale features form the backbone for the subsequent attention-guided, coarse-to-fine registration process.

\subsection{Correspondence-Aware Deformation Propagation}

The decoder performs $L$ stages of coarse-to-fine registration with ACorD Blocks.  
At the coarsest scale $L$, features $F_m^{(L)}$ and $F_f^{(L)}$ are processed by an ACorD block consisting of  
(1) Correlation-guided Cross-scale Offset Prediction (CCOP) and  
(2) Multi-Head Deformable Aggregation (MHDA),  
yielding the initial deformation field $\phi^{(L)}$.
At each finer scale $l < L$, the context feature $F_c^{(l+1)}$ from the previous stage, the warped moving feature $F_w^{(l)} = F_m^{(l)} \circ \phi^{(l+1)}$, and the fixed feature $F_f^{(l)}$ are fed into the CCOP module to predict multi-head offsets $\{\Delta_h^{(l)}\}_{h=1}^{\mathbb{H}_l}$, where $\mathbb{H}_l$ is the number of heads at scale $l$.  
The MHDA module then performs deformable attention at these offset locations, with attention weights $\text{Attention}_h^{(l)}$ 
derived from query–key similarity, so that candidates better matching the fixed feature contribute more to the aggregated coordinate. 
The attended deformation patterns $A_h^{(l)}$ from all heads are regressed to produce the refinement $\Delta \phi^{(l)}$, which is composed with the previous deformation:
$\phi^{(l)} = \phi^{(l+1)} \circ \Delta \phi^{(l)}$.
This process continues to the finest scale, yielding the final deformation field $\phi^{(1)}$.

\subsection{Adaptive Correspondence Discovery Block}
Fig. \ref{fig:fig2} illustrates the core innovation of our framework: Adaptive Correspondence Discovery (ACorD) mechanism, which uses CCOP and MHDA for \textit{attention-guided correspondence discovery}.

\subsubsection{Correlation-guided Cross-scale Offset Prediction:}
CCOP predicts correspondence candidates by estimating offsets relative to a reference neighborhood
 $\mathcal{N}_K(i,j,k) = \{[i + \delta_x, j+ \delta_y, k+\delta_z]: \delta_{xyz} \in [-\lfloor K/2 \rfloor, \lfloor K/2 \rfloor]^3\}$ where $K$ is the search window size. Learning offsets allows the network to 
 adaptively shift the search region toward locations with stronger feature correspondence,
 leveraging both local similarity and cross-scale motion context.

\textbf{Feature Projection and Correlation Computation:} At scale $l$, CCOP first applies shared projection layers to decouple feature extraction from deformation prediction. The fixed and warped moving image features are projected into a correspondence space using a shared MLP and layer normalization:
\begin{equation}
    P_f^{(l)} = \text{Proj}(F_f^{(l)}), \quad P_m^{(l)} = \text{Proj}(F_w^{(l)})
\end{equation}
where $F_w^{(l)}=F_m^{(l)}\circ\phi^{(l+1)} $ and $\phi^{(l+1)}$ is the accumulated deformation from previous scale $l+1$. The number of channels for the projected feature is $\mathbb{H}_l \times C_{head}$, where $\mathbb{H}_l$ is the number of the heads for scale $l$ and $C_{head}$ represents the single-head dimension. The multi-head mechanism will be further introduced in Sec. \ref{MHDA} to capture various deformation pattern for robust registration.

Next, a local correlation volume $C^{(l)} \in \mathbb{R}^{H_l \times W_l \times D_l \times L^3}$ is computed within a cubic neighborhood window of size $L \times L \times L$ to encode spatial similarity prior between the projected fixed and moving features~\cite{kang2022dual,meng2024correlation}. For each voxel location $(i,j,k)$, the correlation with its neighboring voxel $(r_x,r_y,r_z) \in \mathcal{N}_L(i,j,k)$ is given by
\begin{equation}
\small
    C^{(l)}[i,j,k,r] = \text{mean}\!\left(P_f^{(l)}[i,j,k] \odot P_m^{(l)}[r_x,r_y,r_z]\right),
\end{equation}
where $\odot$ denotes element-wise multiplication. This correlation volume provides the spatial prior necessary for the subsequent offset prediction step.

\textbf{Adaptive Offset Prediction:} 
To integrate coarse-level global motion priors that could guide finer-scale refinement, the context feature $F_c^{(l+1)}$ from the previous scale is first upsampled and then concatenated with the current-scale projected features and the correlation volume:
\begin{equation}
   F_{CCOP}^{(l)} = \text{Cat}([P_f^{(l)}, P_m^{(l)}, \text{Up}(F_c^{(l+1)}), C^{(l)}])
\end{equation}

The concatenated features are processed through a lightweight regression network consisting of three convolutional layers, batch normalization, and ReLU activation to jointly predict multi-head offsets $\{\Delta_h^{(l)}\}_{h=1}^{\mathbb{H}_l}$ and updated context features $F_c^{(l)}$:
\begin{equation}
   \{\Delta_h^{(l)}\}_{h=1}^{\mathbb{H}_l},\, F_c^{(l)} = \text{OffsetPredictor}(F_{CCOP}^{(l)}),
\end{equation}
where $\mathbb{H}_l$ denotes the number of attention heads, and each $\Delta_h^{(l)} \in \mathbb{R}^{H_l \times W_l \times D_l \times (K \times K \times K) \times 3}$ encodes the predicted 3D offset $(\Delta x, \Delta y, \Delta z)$ of $K^3$ search region for each voxel in head $h$. The updated context feature $F_c^{(l)}$ is then propagated to the next finer scale to provide hierarchical motion context.
The predicted offsets allows the network to dynamically adapt the search range according to deformation complexity, thus guide the subsequent MHDA module to capture global motion and fine-grained positioning for local refinements within a unified framework

\subsubsection{Multi-Head Deformable Aggregation}
\label{MHDA}

MHDA performs dynamic correspondence aggregation by positioning attention keys at the updated neighbourhood location using the offset predicted by CCOP. Unlike previous fixed-grid attention, MHDA enables flexible correspondence discovery with sparse attention mechanism.

\textbf{Dynamic Key Positioning:}  
For each attention head $h \in \{1, \ldots, \mathbb{H}_l\}$ at scale $l$ and each query location $(i,j,k)$, the MHDA module samples $K^3$ keys from the projected warped moving features at dynamically updated locations using the predicted offsets from the CCOP module. The queries are derived from the projected fixed features, while keys are sampled from offset locations, and the values encode the spatial coordinates of these positions:  
\begin{equation}
\small
\label{eq:q}
    Q_h^{(l)}[i,j,k] = P_f^{(l)}[i,j,k]_h,
\end{equation}
\begin{equation}
\small
\label{eq:k}
    K_h^{(l)}[i,j,k,r] = \text{Sample}\bigl({P_m^{(l)}}_h,\, \mathcal{N}_K(i,j,k)[r] + \Delta_h^{(l)}[i,j,k,r]\bigr),
\end{equation}
\begin{equation}
\small
\label{eq:v}
    V_h^{(l)}[i,j,k,r] = \mathcal{N}_K(i,j,k)[r] + \Delta_h^{(l)}[i,j,k,r],
\end{equation}
where ${P_m^{(l)}}_h$ and ${P_f^{(l)}}_h$ denote the $C_{head}$-dimensional projected features for head $h$. The indices $i \in \{1, \ldots, H_l\}$, $j \in \{1, \ldots, W_l\}$, and $k \in \{1, \ldots, D_l\}$ specify the query voxel, while $r \in \{1, \ldots, K^3\}$ enumerates the $K^3$ sampled positions. The function $\text{Sample}(\cdot,\cdot)$ performs trilinear interpolation, and $V_h^{(l)}[i,j,k,r]$ stores the spatial coordinates of the sampled key positions for head $h$.

\textbf{Attention-Based Correspondence Assessment:} Let $d_k$ be the key dimension. For each query position and attention head, the attention mechanism computes correspondence weights across the $K^3$ key positions:
\begin{equation}
\small
\text{Attention}_h^{(l)}[i, j, k] = \text{Softmax}\left(\frac{Q_h^{(l)}[i, j, k] \cdot K_h^{(l)}[i, j, k, :]}{\sqrt{d_k}}\right)
\end{equation}
The attended coordinate for each head is then computed as:
\begin{equation}
\small
   A_h^{(l)}[i, j, k] = \sum_{r=1}^{K^3} \text{Attention}_h^{(l)}[i,j, k, r] \cdot V_h^{(l)}[i,j, k, r]
\end{equation}

\textbf{Multi-Head Coordinate Aggregation:} For each query position ($i, j, k$), we obtain $\mathbb{H}_l$ attended coordinates from all attention heads. These coordinates are concatenated and processed through a simple regression head to produce the final deformation:
\begin{equation}
\small
   \Delta \phi^{(l)}[i, j, k] = \text{RegHead}(\text{Cat}([A_1^{(l)}[i, j, k], \ldots, A_{\mathbb{H}_l}^{(l)}[i, j, k]]))
\end{equation}

\begin{equation}
    \phi^{(l)} = \phi^{(l+1)}\circ \Delta \phi^{(l)} 
\end{equation}

This multi-head design \cite{ashish2017attention} helps to capture robust correspondence through complementary deformation patterns—each head can specialize in different types of anatomical motion while the concatenated representation provides comprehensive correspondence information. 

\begin{table*}[t]
\centering
\caption{Quantitative comparisons on MindBoggle, LPBA, and OASIS datasets. 
\textcolor{red}{\textbf{Red}}, \textcolor{green}{\textbf{Green}} and \textcolor{blue}{\textbf{Blue}} denote the 
\textcolor{red}{\textbf{1st}}, \textcolor{green}{\textbf{2nd}} and \textcolor{blue}{\textbf{3rd}} best.}
\label{tab:brain_results}
\setlength{\tabcolsep}{7pt}
\renewcommand{\arraystretch}{0.85}
\resizebox{\textwidth}{!}{%
\begin{tabular}{l|l|ccc|ccc|ccc|c}
\toprule
\multirow{2}{*}{\textbf{Group}} & \multirow{2}{*}{\textbf{Method}} 
& \multicolumn{3}{c|}{\textbf{MindBoggle} (97 ROIs)} 
& \multicolumn{3}{c|}{\textbf{LPBA} (54 ROIs)} 
& \multicolumn{3}{c|}{\textbf{OASIS} (35 ROIs)} 
& \multirow{2}{*}{\textbf{\#params (M)}} \\
\cmidrule(lr){3-11}
& & DSC $\uparrow$ & ASSD $\downarrow$ & NJD(\%) $\downarrow$
& DSC $\uparrow$ & ASSD $\downarrow$ & NJD(\%) $\downarrow$
& DSC $\uparrow$ & ASSD $\downarrow$ & NJD(\%) $\downarrow$ & \\
\midrule
\multirow{2}{*}{Traditional} 
& Affine & 0.404$\pm$0.021 & 2.137$\pm$0.196 & \textcolor{red}{\textbf{0}}
& 0.635$\pm$0.020 & 2.120$\pm$0.125 & \textcolor{red}{\textbf{0}}
& 0.571$\pm$0.052 & 1.618$\pm$0.281 & \textcolor{red}{\textbf{0}} & -- \\
& SyN\cite{avants2008symmetric} & 0.513$\pm$0.031 & 1.692$\pm$0.211 & \textcolor{red}{\textbf{0}}
& 0.695$\pm$0.017 & 1.780$\pm$0.109 & \textcolor{red}{\textbf{0}}
& 0.755$\pm$0.029 & 0.837$\pm$0.123 & \textcolor{red}{\textbf{0}} & -- \\
\midrule
\multirow{6}{*}{Single-Scale}
& VoxelMorph\cite{balakrishnan2019voxelmorph} & 0.500$\pm$0.108 & 2.147$\pm$1.092 & 0.883$\pm$0.289
& 0.638$\pm$0.031 & 2.122$\pm$0.242 & 0.589$\pm$0.273
& 0.763$\pm$0.031 & 0.823$\pm$0.145 & 0.963$\pm$0.179 & 0.27 \\
& ViT-V-Net\cite{chen2021vit} & 0.526$\pm$0.112 & 2.086$\pm$1.128 & 0.797$\pm$0.332
& 0.648$\pm$0.030 & 2.049$\pm$0.225 & 0.735$\pm$0.247
& 0.787$\pm$0.031 & 0.733$\pm$0.136 & 0.917$\pm$0.151 & 31.56 \\
& TransMorph\cite{chen2022transmorph} & 0.551$\pm$0.118 & 1.963$\pm$1.083 & 0.691$\pm$0.181
& 0.663$\pm$0.030 & 1.982$\pm$0.232 & 0.648$\pm$0.199
& 0.806$\pm$0.025 & 0.661$\pm$0.103 & 0.827$\pm$0.169 & 46.77 \\
& TransMatch\cite{chen2023transmatch} & 0.544$\pm$0.112 & 1.950$\pm$1.051 & 0.675$\pm$0.200
& 0.667$\pm$0.018 & 1.956$\pm$0.218 & 0.515$\pm$0.162
& 0.805$\pm$0.022 & 0.665$\pm$0.096 & 0.767$\pm$0.138 & 112.41 \\
& TransDCA\cite{chen2023deformable} & 0.550$\pm$0.118 & 1.944$\pm$1.071 & 0.604$\pm$0.161
& 0.661$\pm$0.031 & 1.998$\pm$0.243 & 0.672$\pm$0.240
& 0.803$\pm$0.023 & 0.674$\pm$0.097 & 0.818$\pm$0.166 & 219.67 \\
& H-ViT\cite{ghahremani2024h} & 0.536$\pm$0.113 & 2.033$\pm$1.137 & 0.751$\pm$0.342
& 0.667$\pm$0.030 & 1.969$\pm$0.235 & 0.524$\pm$0.177
& 0.803$\pm$0.021 & 0.676$\pm$0.095 & 0.817$\pm$0.156 & 18.20 \\
\midrule
\multirow{5}{*}{Multi-Scale}
& PIViT\cite{ma2023pivit} & 0.559$\pm$0.075 & 1.737$\pm$0.694 & 0.041$\pm$0.025
& 0.706$\pm$0.013 & 1.712$\pm$0.113 & 0.026$\pm$0.012
& 0.797$\pm$0.022 & 0.684$\pm$0.096 & 0.180$\pm$0.044 & 0.66 \\
& NICE-Net\cite{meng2022non} & 0.589$\pm$0.081 & 1.668$\pm$0.718 & 0.651$\pm$0.239
& 0.715$\pm$0.157 & 1.657$\pm$0.135 & 0.418$\pm$0.107
& 0.821$\pm$0.023 & 0.610$\pm$0.103 & 0.820$\pm$0.160 & 1.08 \\
& NICE-Trans\cite{meng2023non} & 0.608$\pm$0.068 & 1.568$\pm$0.599 & 0.668$\pm$0.231
& 0.726$\pm$0.015 & 1.591$\pm$0.126 & 0.389$\pm$0.093
& \textcolor{green}{\textbf{0.828}$\pm$0.024} & \textcolor{blue}{\textbf{0.588}$\pm$0.097} & 0.822$\pm$0.158 & 5.69 \\
& RPNet\cite{ma2024iirp} & 0.587$\pm$0.076 & 1.659$\pm$0.676 & 0.113$\pm$0.056
& 0.714$\pm$0.015 & 1.675$\pm$0.136 & 0.060$\pm$0.020
& 0.816$\pm$0.023 & 0.624$\pm$0.099 & 0.183$\pm$0.057 & 0.42 \\
& RDPNet\cite{wang2024recursive} & 0.608$\pm$0.0748 & 1.606$\pm$0.659 & \textcolor{green}{\textbf{0.003}$\pm$0.001}
& \textcolor{blue}{\textbf{0.729}$\pm$0.015} & 1.589$\pm$0.124 & \textcolor{green}{\textbf{0.001}$\pm$0.001}
& \textcolor{red}{\textbf{0.830}$\pm$0.023} & \textcolor{green}{\textbf{0.581}$\pm$0.096} & \textcolor{blue}{\textbf{0.008}$\pm$0.003} & 8.92 \\
\midrule
\multirow{5}{*}{Local-Corr.}
& Dual-PR++\cite{kang2022dual} & 0.562$\pm$0.102 & 1.899$\pm$0.974 & 0.258$\pm$0.111
& 0.695$\pm$0.021 & 1.798$\pm$0.183 & 0.040$\pm$0.018
& 0.807$\pm$0.021 & 0.657$\pm$0.095 & 0.449$\pm$0.106 & 1.24 \\
& Im2Grid\cite{liu2022coordinate} & 0.556$\pm$0.098 & 1.768$\pm$0.788 & 0.019$\pm$0.006
& 0.703$\pm$0.015 & 1.715$\pm$0.129 & 0.008$\pm$0.003
& 0.804$\pm$0.021 & 0.666$\pm$0.095 & 0.028$\pm$0.012 & 0.89 \\
& ModeT\cite{wang2023modet} & \textcolor{blue}{\textbf{0.614}$\pm$0.051} & \textcolor{green}{\textbf{1.472}$\pm$0.418} & 0.008$\pm$0.006
& \textcolor{blue}{\textbf{0.729}$\pm$0.013} & \textcolor{blue}{\textbf{1.573}$\pm$0.110} & 0.002$\pm$0.001
& \textcolor{blue}{\textbf{0.824}$\pm$0.023} & 0.600$\pm$0.095 & 0.013$\pm$0.006 & 1.03 \\
& CorrMLP\cite{meng2024correlation} & \textcolor{green}{\textbf{0.617}$\pm$0.057} & 1.533$\pm$0.528 & 0.599$\pm$0.291
& \textcolor{green}{\textbf{0.731}$\pm$0.015} & \textcolor{green}{\textbf{1.566}$\pm$0.110} & 0.307$\pm$0.078
& \textcolor{red}{\textbf{0.830}$\pm$0.023} & \textcolor{red}{\textbf{0.580}$\pm$0.093} & 0.761$\pm$0.143 & 4.20 \\
& SACB-Net\cite{cheng2025sacb} & 0.605$\pm$0.055 & \textcolor{blue}{\textbf{1.514}$\pm$0.455} & 0.125$\pm$0.041
& 0.725$\pm$0.013 & 1.581$\pm$0.108 & 0.103$\pm$0.019
& 0.818$\pm$0.023 & 0.616$\pm$0.095 & 0.208$\pm$0.039 & 1.11 \\
\midrule
Proposed & \textbf{ACorD} 
& \textcolor{red}{\textbf{0.624}$\pm$0.049} & \textcolor{red}{\textbf{1.468}$\pm$0.447} & \textcolor{blue}{\textbf{0.005}$\pm$0.004}
& \textcolor{red}{\textbf{0.736}$\pm$0.015} & \textcolor{red}{\textbf{1.549}$\pm$0.123} & \textcolor{blue}{\textbf{0.002}$\pm$0.001}
& \textcolor{red}{\textbf{0.830}$\pm$0.024} & \textcolor{red}{\textbf{0.580}$\pm$0.099} & \textcolor{green}{\textbf{0.006}$\pm$0.003} 
& \textbf{6.47} \\
\bottomrule
\end{tabular}%
}
\end{table*}

\subsection{Unsupervised Training Loss Function}

Our method is trained with an unsupervised loss combining three components computed on the final deformation field  $\phi$: 
$\mathcal{L} = \mathcal{L}_{\text{sim}} + \lambda_{\text{smooth}} \mathcal{L}_{\text{smooth}} + \lambda_{\text{jac}} \mathcal{L}_{\text{jac}}$.
The similarity term uses normalized cross-correlation (NCC) to measure alignment quality: 
$\mathcal{L}_{\text{sim}} = - NCC(I_f, I_m\circ \phi)$.
Smoothness regularization enforces spatially coherent deformations: 
$\mathcal{L}_{\text{smooth}} = \sum_{d \in \{x,y,z\}} ||\nabla \phi_d^{(1)}||_2^2$.

The Jacobian regularization penalizes negative determinants to prevent folding: 
$\mathcal{L}_{jac}=\frac{1}{N} \sum_{\boldsymbol{p} \in \Omega} \operatorname{relu}\left(-\left|J_\phi(\boldsymbol{p})\right|\right)$. In the above formula $\lambda_{\text{smooth}}$ and $\lambda_{\text{jac}}$
  balance smoothness and topology preservation.

\section{Experiments}

\subsection{Datasets}

We evaluated our framework on three public 3D brain MRI datasets and one abdominal CT dataset.

\textbf{MindBoggle} \cite{klein2017mindboggling} consists of 101 brain MRI scans with 97 anatomical ROIs, including both cortical (62) and subcortical (35) structures. 
We used 70 scans for training, 7 for validation, and a separate set of 24 for cross-site testing to ensure generalizability.

\textbf{LPBA40} \cite{shattuck2008construction} contains 40 healthy adult brain MRI scans with 54 manual anatomical labels. We used 25 scans for training, 5 for validation, and 10 for testing.

\textbf{OASIS} \cite{marcus2010open} contains 414 brain MRI scans with 35 anatomical labels from a cross-sectional cohort spanning young, middle-aged, nondemented, and demented older adults. We used 200 scans for training, 20 for validation, and the provided test pairs for testing.

\textbf{AbdomenCT} \cite{xu2016evaluation} comprises 30 abdominal CT scans with 13 anatomical RoIs. This dataset is particularly challenging due to significant anatomical variations,
making it a challenging testbed for large-deformation registration. We used 18 scans for training, 4 for validation, and 8 for testing.

\subsection{Implementation Details}

Our method was implemented in PyTorch and trained on a single NVIDIA RTX A6000 GPU.
Runtime, MACs, and peak GPU memory are reported in the supplementary material; ACorD takes 710 ms/pair and 16.9 GB peak memory, and is not the lightest model.
We use a 5-scale pyramid and the local correlation neighborhood size is set to $L=3$ to provide the spatial prior for offset prediction, and the dynamic attention search window $K$ is set to 3.
Training was performed using Adam optimizer with a learning rate of $10^{-4}$ and linear annealing. The loss function hyperparameters were set at $\lambda_{\text{smooth}} = 1$ and $\lambda_{\text{jac}} = 50$ after tuning.


\begin{table}[t]
\centering
\caption{Quantitative comparison on AbdomenCT dataset. \textcolor{red}{\textbf{Red}}, \textcolor{green}{\textbf{Green}} and \textcolor{blue}{\textbf{Blue}} denote the \textcolor{red}{\textbf{1st}}, \textcolor{green}{\textbf{2nd}} and \textcolor{blue}{\textbf{3rd}} best resutls.}
\label{tab:abdomen_results}

\resizebox{0.4\textwidth}{!}{
\scriptsize
\begin{tabular}{l@{\hspace{2pt}}|l@{\hspace{2pt}}|c@{\hspace{4pt}}c@{\hspace{4pt}}c@{\hspace{1pt}}}
\toprule
\multirow{2}{*}{\textbf{Group}} & \multirow{2}{*}{\textbf{Method}} & \multicolumn{3}{c}{\textbf{AbdomenCT} (13 ROIs)} \\
\cmidrule(lr){3-5}
& & DSC $\uparrow$ & ASSD $\downarrow$ & NJD(\%) $\downarrow$ \\
\midrule
\multirow{2}{*}{Traditional} 
& Affine & 0.251$\pm$0.070 & 6.629$\pm$2.416 & \textcolor{red}{\textbf{0}} \\
& SyN\cite{avants2008symmetric} & 0.319$\pm$0.084 & 5.780$\pm$2.443 & \textcolor{red}{\textbf{0}} \\
\midrule
\multirow{5}{*}{Single-Scale}
& VoxelMorph\cite{balakrishnan2019voxelmorph} & 0.349$\pm$0.051 & 4.793$\pm$1.047 & 1.746$\pm$0.094 \\
& ViT-V-Net\cite{chen2021vit} & 0.385$\pm$0.054 & 4.547$\pm$1.032 & 4.798$\pm$1.400 \\
& TransMorph\cite{chen2022transmorph} & 0.409$\pm$0.061 & 4.422$\pm$1.118 & 5.554$\pm$1.368 \\
& TransMatch\cite{chen2023transmatch} & 0.420$\pm$0.064 & 4.258$\pm$1.097 & 5.188$\pm$1.430 \\
& TransDCA\cite{chen2023deformable} & 0.420$\pm$0.058 & 4.317$\pm$1.071 & 4.497$\pm$1.375 \\
& H-ViT\cite{ghahremani2024h} & 0.429$\pm$0.055 & 4.147$\pm$0.986 & 5.219$\pm$2.250 \\
\midrule
\multirow{5}{*}{Multi-Scale}
& PIViT\cite{ma2023pivit} & 0.416$\pm$0.044 & 4.005$\pm$0.714 & 0.367$\pm$0.179 \\
& NICE-Net\cite{meng2022non} & 0.464$\pm$0.058 & 3.666$\pm$0.893 & 6.375$\pm$2.100 \\
& NICE-Trans\cite{meng2023non} & 0.524$\pm$0.047 & 3.034$\pm$0.689 & 14.938$\pm$2.943 \\
& RPNet\cite{ma2024iirp} & 0.480$\pm$0.055 & 3.543$\pm$0.723 & 1.700$\pm$0.827 \\
& RDPNet\cite{wang2024recursive} & 0.554$\pm$0.052 & 3.023$\pm$0.757 & 0.227$\pm$0.216 \\
\midrule
\multirow{4}{*}{Local-Corr.}
& Dual-PR++\cite{kang2022dual} & 0.466$\pm$0.056 & 3.883$\pm$0.929 & 2.371$\pm$1.208 \\
& Im2Grid\cite{liu2022coordinate} & 0.455$\pm$0.041 & 3.642$\pm$0.686 & 0.178$\pm$0.119 \\
& ModeT\cite{wang2023modet} & 0.554$\pm$0.053 & 2.729$\pm$0.646 & \textcolor{green}{\textbf{0.061}$\pm$0.029} \\
& CorrMLP\cite{meng2024correlation} & \textcolor{blue}{\textbf{0.572}$\pm$0.055} & \textcolor{blue}{\textbf{2.645}$\pm$0.708} & 12.235$\pm$2.686 \\
& SACB-Net\cite{cheng2025sacb} & \textcolor{green}{\textbf{0.577}$\pm$0.047} & \textcolor{green}{\textbf{2.575}$\pm$0.603} & 1.642$\pm$0.651 \\
\midrule
Proposed & \textbf{ACorD} & \textcolor{red}{\textbf{0.597}$\pm$0.052} & \textcolor{red}{\textbf{2.411}$\pm$0.603} & \textcolor{blue}{\textbf{0.136}$\pm$0.092} \\
\bottomrule
\end{tabular}
}
\end{table}

\subsection{Evaluation Metrics}

We evaluate registration accuracy using the \textbf{Dice Similarity Coefficient (DSC)}, measuring anatomical overlap, and \textbf{Average Symmetric Surface Distance (ASSD)} \cite{taha2015metrics, heimann2009comparison} quantifies surface alignment accuracy. The smoothness and invertibility of the predicted spatial transformations are evaluated using the percentage of \textbf{Negative Jacobian Determinants (NJD)} \cite{dalca2019unsupervised}, indicating topological violations.

\subsection{Comparison Methods}


We compared ACorD with state-of-the-art registration methods, including \textbf{2} traditional optimization-based methods and \textbf{16} deep registration methods grouped into three categories (as discussed in Sec.\ref{related_works}): \textbf{Single-Scale} methods ~\cite{balakrishnan2019voxelmorph, chen2021vit, chen2022transmorph, chen2023transmatch, chen2023deformable, ghahremani2024h}; \textbf{Multi-Scale} methods ~\cite{ma2023pivit, meng2022non,meng2023non, ma2024iirp, wang2024recursive}; and \textbf{Local Correlation} methods ~\cite{kang2022dual,liu2022coordinate, wang2023modet,meng2024correlation,cheng2025sacb}. 
All methods were retrained using the same data splits and preprocessing pipelines. To preserve each method's
intended formulation, we follow its original training objective and regularization settings.

\section{Results and Discussion}

\subsection{Brain Registration Results}

\begin{figure*}[t]
  \centering
  \includegraphics[width=\linewidth]{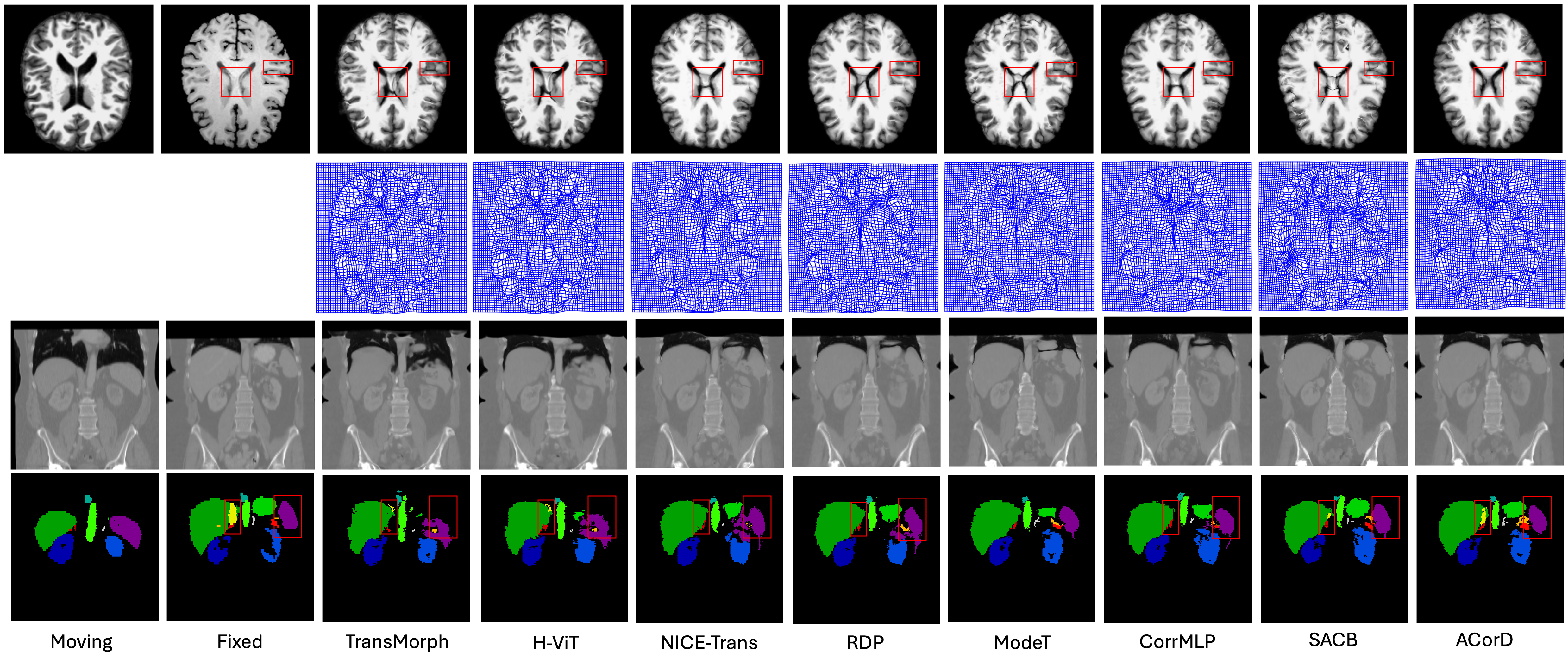}

   \caption{Visual comparisons on the Brain LPBA dataset (top two rows) and the Abdomen CT dataset (bottom two rows)}
   \label{fig:lpba_vis}
\end{figure*}

Table~\ref{tab:brain_results} presents evaluations across three brain MRI datasets. \textbf{ACorD} consistently demonstrates superior performance across all evaluation metrics, including DSC, ASSD, and the percentage NJD.
Single-scale methods generally underperform multi-scale and correlation-based approaches. Among multi-scale methods RDPNet achieves strong performance through recursive multi-step refinement and maintains excellent topology preservation via diffeomorphic registration. Local-window correlation methods show promising results, with ModeT and CorrMLP achieving competitive performance, though still constrained to fixed neighborhoods. 
Our method outperforms all methods across the three datasets, achieving the highest DSC and ASSD scores on MindBoggle and LPBA, while matching the best performance on OASIS. 
Notably, these improvements are achieved with a moderate model size (6.47M parameters), which is substantially smaller than transformer-based methods such as TransDCA, demonstrating that adaptive correspondence discovery and selective aggregation can improve accuracy without excessive model complexity.
Fig.~\ref{fig:lpba_vis} shows qualitative LPBA results with improved anatomical alignment and fewer structural distortions.

\begin{figure}[h]
\centerline{\includegraphics[width=\columnwidth]{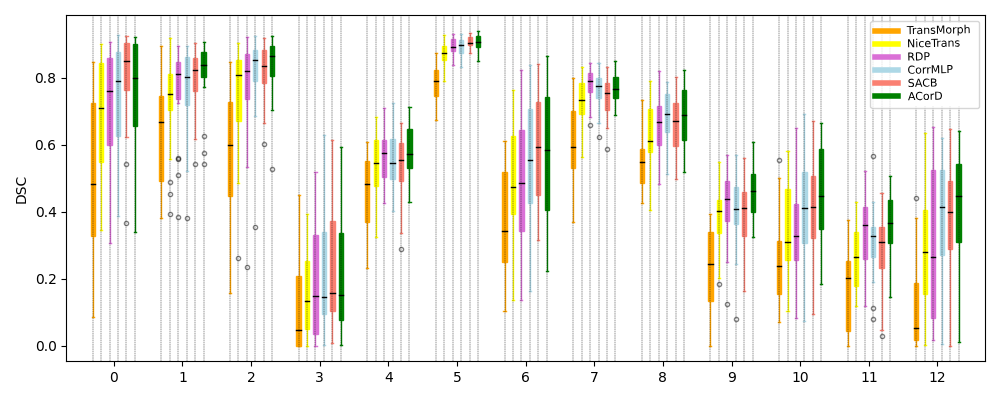}}
\caption{Dice score evaluation on AbdomenCT (13 ROIs)}
\label{fig:box_plot_abdomenct}
\end{figure}
\subsection{AbdomenCT Registration Results}
AbdomenCT is challenging due to large, non-linear organ deformations from breathing and positional changes. As such, it serves as a critical testbed for evaluating a registration model's robustness to large displacements. As presented in Table~\ref{tab:abdomen_results}, ACorD achieves state-of-the-art performance on this challenging dataset.

Compared to brain datasets, performance differences between method categories become more pronounced, emphasizing the importance of adaptive correspondence modeling under large deformations. 
SACB-Net is the strongest competitor, demonstrating the effectiveness of explicit structure modeling and correspondence mechanisms. However, its fixed neighborhood constraint limits effectiveness for the large displacements. By contrast, ACorD benefits from adaptive, context-conditioned correspondence discovery, enabling more effective alignment of distant structures while maintaining reasonable topology preservation.
 The results underscore that attention-guided propagation becomes increasingly valuable as deformation complexity increases. As further illustrated in Fig.~\ref{fig:box_plot_abdomenct}, ACorD consistently achieves higher Dice scores across abdominal structures, confirming its robustness under large deformations.
 
\textbf{Effect of Jacobian regularization.}
We additionally tuned $\lambda_{\mathrm{jac}}$ for the strongest competitors. CorrMLP reaches a best DSC of $0.580$ and ModeT reaches $0.565$, while SACB-Net does not improve over its original configuration. ACorD achieves $0.597$ DSC under its selected setting. These results indicate that ACorD's improvement cannot be explained solely by stronger Jacobian regularization.

\subsection{Ablation Study}


\begin{table}[t]
\centering
\caption{Ablation study on the AbdomenCT dataset.
\textbf{CG}: Correlation Guidance;
\textbf{CS}: Cross-Scale Feature Propagation;
\textbf{DA}: Deformation Aggregation;
\textbf{MH}: Multi-Head mechanism;
\textbf{Agg}: aggregation rule
(\textbf{Attn}: attention-weighted, \textbf{Avg}: uniform averaging);
\textbf{Offset}: correspondence offset source
(\textbf{Learned}: CCOP-predicted, \textbf{Default}: fixed neighbors).}
\label{tab:ablation}
\resizebox{0.9\columnwidth}{!}{
\small
\begin{tabular}{c| c c c c c c | c c c}
\toprule
\multicolumn{1}{c|}{} 
& \multicolumn{6}{c|}{\textbf{Components}} 
& \multicolumn{3}{c}{\textbf{AbdomenCT (13 ROIs)}} \\
\midrule
Exp & CG & CS & DA & MH & Agg & Offset 
& DSC $\uparrow$ & ASSD $\downarrow$ & NJD(\%) $\downarrow$ \\
\midrule
1 &  &  &  &  &  &  
& 0.549$\pm$0.054 & 2.83$\pm$0.67 & 0.105$\pm$0.08 \\
2 & \checkmark &  &  &  &  &  
& 0.570$\pm$0.051 & 2.731$\pm$0.658 & 0.103$\pm$0.066 \\
3 & \checkmark & \checkmark &  &  &  &  
& 0.577$\pm$0.046 & 2.626$\pm$0.646 & 0.151$\pm$0.105 \\
4 &  &  & \checkmark &  & Attn & Learned 
& 0.564$\pm$0.045 & 2.649$\pm$0.598 & 0.098$\pm$0.064 \\
5 &  &  & \checkmark & \checkmark & Attn & Learned 
& 0.579$\pm$0.049 & 2.585$\pm$0.651 & 0.084$\pm$0.048 \\
6 &  &  & \checkmark & \checkmark & Avg & Learned 
& 0.557$\pm$0.048 & 2.686$\pm$0.590 & 0.048$\pm$0.029 \\
7 & \checkmark & \checkmark & \checkmark & \checkmark & Avg & Learned 
& 0.583$\pm$0.052 & 2.543$\pm$0.650 & 0.089$\pm$0.066 \\
8 & \checkmark & \checkmark & \checkmark & \checkmark & Attn & Default 
& 0.559$\pm$0.053 & 2.657$\pm$0.612 & 0.014$\pm$0.092 \\
\midrule
\textbf{ACorD} & \checkmark & \checkmark & \checkmark & \checkmark & Attn & Learned 
& \textbf{0.597$\pm$0.052} & \textbf{2.411$\pm$0.603} & \textbf{0.136$\pm$0.092} \\
\bottomrule
\end{tabular}
}
\end{table}

\begin{table}[t]
\centering
\caption{Aggregated position label accuracy (\%) using uniform and attention-based aggregation over ACorD candidates.}
\label{tab:aggregation_accuracy}

\scriptsize
\setlength{\tabcolsep}{8pt}
\renewcommand{\arraystretch}{0.65}

\begin{tabular}{ccc|cc}
\toprule
\multirow{2}{*}{\textbf{Lvl.}}
& \multicolumn{2}{c}{\textbf{MindBoggle}}
& \multicolumn{2}{c}{\textbf{AbdomenCT}} \\
\cmidrule(lr){2-3}
\cmidrule(lr){4-5}
& \textbf{Unif.} & \textbf{Attn.}
& \textbf{Unif.} & \textbf{Attn.} \\
\midrule
1 & 73.89 & \textbf{74.40} & 78.35 & \textbf{78.56} \\
2 & 65.95 & \textbf{68.07} & 74.78 & \textbf{75.84} \\
3 & 55.55 & \textbf{58.03} & 68.79 & \textbf{70.79} \\
4 & 49.56 & \textbf{49.64} & 59.14 & \textbf{62.07} \\
5 & \textbf{49.19} & 48.91 & \textbf{48.29} & 47.31 \\
\bottomrule
\end{tabular}
\end{table}

\subsubsection{Components Analysis}
Table~\ref{tab:ablation} presents an ablation study that systematically evaluates the contribution of each component in the proposed framework. 
Correlation guidance (CG) improves DSC and ASSD by providing explicit correspondence cues, while cross-scale feature propagation (CS) adds global deformation context for more informed refinement (Exp. ~2 and ~3).
The deformable aggregation module (DA) improves topology preservation, and its multi-head extension (MH) further enhances performance by capturing diverse correspondence hypotheses (Exp.~4 and ~5). 

To assess the role of learned candidate weighting, we replace attention-weighted aggregation with uniform averaging while keeping all other components unchanged. As shown in Exp.~6, this substitution degrades performance, indicating that selectively weighting correspondence candidates is critical for effective deformation refinement. 
This is further supported by comparing Exp.~7 with the full ACorD model, further confirming that the performance gain arises from learned candidate weighting rather than generating multiple candidates.

We additionally evaluate adaptive correspondence discovery by replacing learned offsets with fixed local neighbors while retaining attention-based aggregation (Exp.~8). The performance drop confirms that attention alone is insufficient, and adaptive, context-conditioned offset prediction is essential for discovering informative correspondences across varying deformation scales.

Finally, the complete ACorD framework, which integrates correlation guidance, cross-scale context, adaptive correspondence discovery, and learned candidate weighting, achieves the best overall performance across all metrics. These results validate the effectiveness of selectively aggregating adaptively discovered correspondences for robust multi-scale medical image registration.

Beyond the end-task ablation, we verify within a single trained model that replacing uniform averaging with attention weighting over an identical candidate set increases the anatomical correctness of the aggregated position at 4 of 5 scales on both AbdomenCT and MindBoggle. (Table \ref{tab:aggregation_accuracy}).

\begin{table}[t]
\centering
\caption{Common-set recovery on AbdomenCT. For voxels misregistered by both models at the coarser scale, we report the percentage corrected one level finer. Higher is better; p-values use paired Wilcoxon signed-rank tests.}
\label{tab:common_set_recovery}
{
\scriptsize
\setlength{\tabcolsep}{3pt}
\renewcommand{\arraystretch}{0.65}
\begin{tabular}{c|cc|cc}
\toprule
\textbf{Transition}
& \textbf{ACorD(\%)}
& \textbf{Exp.7(\%)}
& \textbf{ACorD(\%)}
& \textbf{Exp.8(\%)} \\
\midrule
L4$\rightarrow$L3
& \textbf{40.4}
& 38.3 ($p=0.004$)
& \textbf{40.2}
& 37.0 ($p<1\mathrm{e}{-4}$) \\

L3$\rightarrow$L2
& \textbf{23.6}
& 22.6 ($p=0.032$)
& \textbf{23.8}
& 20.6 ($p<1\mathrm{e}{-4}$) \\

L2$\rightarrow$L1
& \textbf{13.2}
& 12.2 ($p=0.001$)
& \textbf{12.6}
& 10.4 ($p<1\mathrm{e}{-4}$) \\
\bottomrule
\end{tabular}
}
\vspace{-5px}
\end{table}
\subsubsection{Cross-Scale Error Recovery}
We next evaluate registration at each pyramid level using its predicted deformation field. On AbdomenCT, ACorD achieves DSC values of 0.304, 0.408, 0.500, 0.563, and 0.597 from \(L_5\) to \(L_1\), exceeding both Learned+Avg and Default+Attn at each level. To distinguish improved recovery from a stronger predictor, we define a common-error set for each coarse-to-fine transition: voxels that are misregistered by both compared models at the coarser level. We then measure the fraction of these shared errors corrected at the next finer level. ACorD recovers a larger fraction at every evaluated transition (Table \ref{tab:common_set_recovery}). We omit \(L_5\to L_4\) because too few common errors remain at \(1/16\) resolution for meaningful significance testing. These results provide direct evidence that ACorD more effectively corrects shared coarse-scale failures during refinement.



\begin{table}[t]
\centering
\caption{Statistics of predicted candidate offsets. \textbf{Ratio}: the ratio of predicted offsets to default offset magnitude. \textbf{Frac. $>1$:} fraction of predicted candidate offsets with ratios greater than 1.}
\label{tab:offset_stats}
{\tiny
\setlength{\tabcolsep}{1.5pt}
\renewcommand{\arraystretch}{0.85}
\begin{tabular}{c|cc|cc|cc|cc}
\toprule
\multirow{2}{*}{\textbf{Lvl.}}
& \multicolumn{2}{c|}{\textbf{AbdomenCT}}
& \multicolumn{2}{c|}{\textbf{LPBA}}
& \multicolumn{2}{c|}{\textbf{OASIS}}
& \multicolumn{2}{c}{\textbf{MindBoggle}} \\
\cmidrule(lr){2-9}
& \textbf{Ratio} & \textbf{Frac. $>1$}
& \textbf{Ratio} & \textbf{Frac. $>1$}
& \textbf{Ratio} & \textbf{Frac. $>1$}
& \textbf{Ratio} & \textbf{Frac. $>1$} \\
\midrule
1 & 1.46$\pm$0.02 & 0.73$\pm$0.01
  & 1.45$\pm$0.07 & 0.75$\pm$0.03
  & 1.07$\pm$0.01 & 0.55$\pm$0.00
  & 1.05$\pm$0.02 & 0.51$\pm$0.05 \\
2 & 1.14$\pm$0.02 & 0.55$\pm$0.01
  & 1.18$\pm$0.03 & 0.68$\pm$0.02
  & 0.68$\pm$0.00 & 0.11$\pm$0.00
  & 1.09$\pm$0.02 & 0.58$\pm$0.01 \\
3 & 0.93$\pm$0.02 & 0.35$\pm$0.02
  & 0.95$\pm$0.01 & 0.32$\pm$0.01
  & 0.79$\pm$0.01 & 0.20$\pm$0.00
  & 0.97$\pm$0.01 & 0.36$\pm$0.01 \\
4 & 1.15$\pm$0.03 & 0.50$\pm$0.02
  & 0.65$\pm$0.01 & 0.09$\pm$0.01
  & 0.75$\pm$0.00 & 0.20$\pm$0.00
  & 0.65$\pm$0.03 & 0.13$\pm$0.03 \\
5 & 1.54$\pm$0.04 & 0.73$\pm$0.01
  & 1.10$\pm$0.01 & 0.49$\pm$0.01
  & 0.97$\pm$0.01 & 0.29$\pm$0.01
  & 0.91$\pm$0.00 & 0.19$\pm$0.00 \\
\bottomrule
\end{tabular}
}
\end{table}
\subsubsection{Statistical Analysis on Predicted Offsets}
Table~\ref{tab:offset_stats} compares ACorD's predicted candidate offsets with the default local offsets across datasets and network levels. AbdomenCT shows the largest and most consistent offset expansion, with predicted offsets exceeding the default magnitude across most levels. This supports the role of adaptive offset prediction for large-deformation registration, where fixed local neighborhoods may be insufficient. In contrast, the brain datasets generally show smaller ratios, suggesting that ACorD adjusts its search range according to deformation complexity.


\subsection{Limitations}
A key direction for future work is improving structure and boundary awareness during training. While ACorD provides effective correspondence discovery, mismatched regions can still occur (see supplementary material). These errors often arise from structure-unaware loss functions that prioritize image similarity, leading to unnecessary or excessive deformation that ignores structural integrity. For example, brain registration results can sometimes show extra deformation that causes structural folding to achieve higher image similarity. Similarly, in AbdomenCT registration, even when images appear visually similar, mask visualizations may reveal structure-level artifacts. Although methods like SACB-Net\cite{cheng2025sacb} have attempted to address this with clustering or graph, they often prove ineffective for complex structures like the brain. 
Therefore, a significant challenge remains: developing a method for structure-aware training that works effectively with non-local deformations. A promising direction is to explore incorporating learned anatomical priors into the attention mechanism, allowing the model to weigh correspondences not just by feature similarity but also by their structural context.

\section{Conclusion}

We presented Adaptive Correspondence Discovery (ACorD), a novel framework for multi-scale medical image registration that enables attention-guided propagation and adaptive correspondence discovery beyond fixed local windows. By predicting correspondence candidates and applying sparse attention adaptively, ACorD achieves accurate alignment across a wide range of deformations. Extensive experiments on four benchmark datasets demonstrate state-of-the-art performance, particularly in handling large non-linear deformations. In future work, we plan to incorporate structure-aware supervision to enhance boundary-level accuracy, extend ACorD to multi-modal registration tasks, and improve scalability.

{
    \small

}



\clearpage

\twocolumn[{
\centering
{\LARGE\bfseries Supplementary Material\par}
\vspace{0.5em}
{\large\bfseries ACorD: Adaptive Correspondence Discovery
for Medical Image Registration\par}
\vspace{1em}
}]

\setcounter{section}{0}
\renewcommand{\thesection}{\Alph{section}}

\setcounter{figure}{0}
\renewcommand{\thefigure}{S\arabic{figure}}

\setcounter{table}{0}
\renewcommand{\thetable}{S\arabic{table}}

\setcounter{equation}{0}
\renewcommand{\theequation}{S\arabic{equation}}

\begin{table*}[h]
\centering
\caption{Deformable Attention Size}
\label{tab:supp}
\resizebox{0.7\textwidth}{!}{
\begin{tabular}{cccc|ccc}
\toprule
\multirow{2}{*}{$K^3$} & \multicolumn{3}{c|}{\textbf{AbdomenCT}} & \multicolumn{3}{c}{\textbf{MindBoggle}} \\
\cmidrule(lr){2-7}
 & DSC$\uparrow$ & ASSD $\downarrow$ & NJD(\%)$\downarrow$ & DSC$\uparrow$ & ASSD $\downarrow$ & NJD(\%)$\downarrow$ \\ 
 \midrule
$2^3=$8 & 0.595$\pm$0.056 & 2.475$\pm$0.614 & 0.135$\pm$0.096 & 0.621$\pm$0.054 & 1.512$\pm$0.517 & 0.005$\pm$0.006 \\

$3^3=27$ & 0.597$\pm$0.052 & 2.411$\pm$0.603 & 0.136$\pm$0.092 & 0.624$\pm$0.049  & 1.468$\pm$0.447 & 0.005$\pm$0.004\\
\bottomrule
\end{tabular}
}
\end{table*}
\section{Ablation Study on Deformable Attention Size $K^3$ }

Based on Table \ref{tab:supp}, we investigated the impact of the attention window size on our model's performance. We specifically evaluated small window sizes of $K=2$ and $K=3$, which both yielded strong results compared with baseline with no attention involved in Table. 3 in the orginal paper. While larger values of $K$ could potentially be explored, we refrained from doing so due to the cubic increase in computational cost and memory requirements. We concluded that our chosen configuration strikes a balance between performance and computational efficiency.

\section{Network Structure for the Encoder layers and Offset Predictor.}
\begin{figure}[h]
\centering
\includegraphics[width=0.7\linewidth]{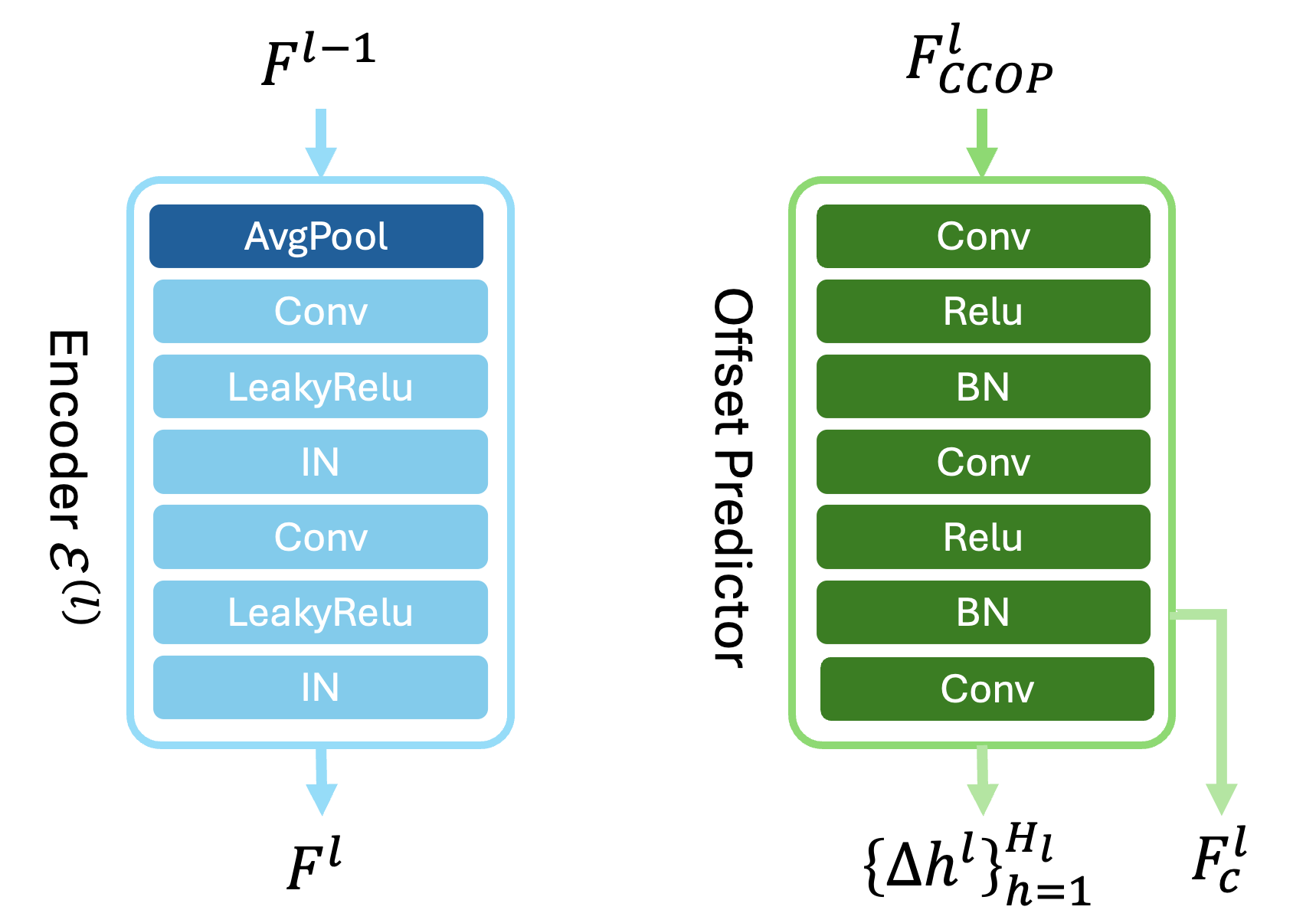}

\caption{Network Structure for the Encoder layers and Offset Predictor.}
\label{fig:supp}
\end{figure}

\section{Details on Loss function}
\subsection{NCC}
\begin{equation}
\begin{aligned}
& N C C\left(I_f, I_w\right)= \\
& \sum_{\boldsymbol{p} \in \Omega} \frac{\left(\sum_{\boldsymbol{p}_i}\left(I_w\left(\boldsymbol{p}_i\right)-\bar{I}_w(\boldsymbol{p})\right)\left(I_f\left(\boldsymbol{p}_i\right)-\bar{I}_f(\boldsymbol{p})\right)\right)^2}{\sum_{\boldsymbol{p}_i}\left(I_w\left(\boldsymbol{p}_i\right)-\bar{I}_w(\boldsymbol{p})\right)^2 \sum_{\boldsymbol{p}_i}\left(I_f\left(\boldsymbol{p}_i\right)-\bar{I}_f(\boldsymbol{p})\right)^2}
\end{aligned}
\end{equation}

\subsection{Jacobian Matrix}
\begin{equation}
J_\phi(\boldsymbol{p})=\left[\begin{array}{lll}
\frac{\partial \phi_x(\boldsymbol{p})}{\partial x} & \frac{\partial \phi_x(\boldsymbol{p})}{\partial y} & \frac{\partial \phi_x(\boldsymbol{p})}{\partial z} \\
\frac{\partial \phi_y(\boldsymbol{p})}{\partial x} & \frac{\partial \phi_y(\boldsymbol{p})}{\partial y} & \frac{\partial \phi_y(\boldsymbol{p})}{\partial z} \\
\frac{\partial \phi_z(\boldsymbol{p})}{\partial x} & \frac{\partial \phi_z(\boldsymbol{p})}{\partial y} & \frac{\partial \phi_z(\boldsymbol{p})}{\partial z}
\end{array}\right]
\end{equation}

\section{Deformation Accuracy with $\lambda_{jac}$ Tuning}
\begin{table}[h]
\centering
\caption{Hyperparameter Tuning for Negative Jacobian Loss}
\label{tab:ablation2}
\resizebox{0.45\textwidth}{!}{
\scriptsize
\begin{tabular}{c|p{1cm}p{1.5cm}|p{1cm}p{1.5cm}}
\toprule
\multirow{3}{*}{$\lambda_{jac}$} & \multicolumn{4}{c}{\textbf{ACorD}} \\
\cmidrule(lr){2-5}
& \multicolumn{2}{c|}{\textbf{AbdomenCT}} & \multicolumn{2}{c}{\textbf{MindBoggle}} \\
\cmidrule(lr){2-5}
& DSC$\uparrow$ & NJD(\%)$\downarrow$ & DSC$\uparrow$ & NJD(\%)$\downarrow$ \\
\midrule
0  & 0.590 & 8.381 & 0.625 & 0.191 \\
20 & 0.592 & 0.317 & 0.623 & 0.011 \\
50 & 0.597 & 0.136 & 0.624 & 0.005 \\
80 & 0.595 & 0.086 & 0.623 & 0.003 \\
\bottomrule
\end{tabular}
}
\end{table}

Table~\ref{tab:ablation2} presents a parameter tuning analysis for the Jacobian determinant loss, $L_{jac}$, a regularization term often overlooked in previous works. Increasing $\lambda_{jac}$ consistently reduces NJD(\%), indicating improved topology preservation after deformation. While DSC remains relatively stable on the MindBoggle brain dataset, moderate Jacobian regularization improves AbdomenCT performance, suggesting that topology preservation is particularly important for large-deformation registration.

\section{Head Allocation Robustness}
\begin{table}[h]
\centering
\caption{Ablation Study on Head Allocation}
\label{tab:head_allocation}
\resizebox{0.45\textwidth}{!}{
\scriptsize
\begin{tabular}{c|c|c|c}
\toprule
\textbf{Head Allocation} & DSC$\uparrow$ & ASSD$\downarrow$ & NJD(\%)$\downarrow$ \\
\midrule
$[2,2,4,4,1]$ & $0.595 \pm 0.057$ & $2.428 \pm 0.601$ & $0.18 \pm 0.14$ \\
$[4,4,2,2,1]$ & $0.594 \pm 0.052$ & $2.418 \pm 0.603$ & $0.094 \pm 0.05$ \\
default $[8,4,2,1,1]$ & $0.597 \pm 0.052$ & $2.411 \pm 0.603$ & $0.136 \pm 0.092$ \\
\bottomrule
\end{tabular}
}
\end{table}

Alternative head allocations on AbdomenCT result in similar performance as presented in Table \ref{tab:head_allocation}: $[2,2,4,4,1]$ achieves 0.595 DSC and $[4,4,2,2,1]$ achieves 0.594 DSC, close to our default $[8,4,2,1,1]$ result of 0.597 DSC, indicating that ACorD is not highly sensitive to the exact head allocation, while the default provides the best accuracy-memory trade-off.

\section{Dice Scores On LPBA dataset}
\begin{figure*}
\centering
\includegraphics[width=\linewidth]{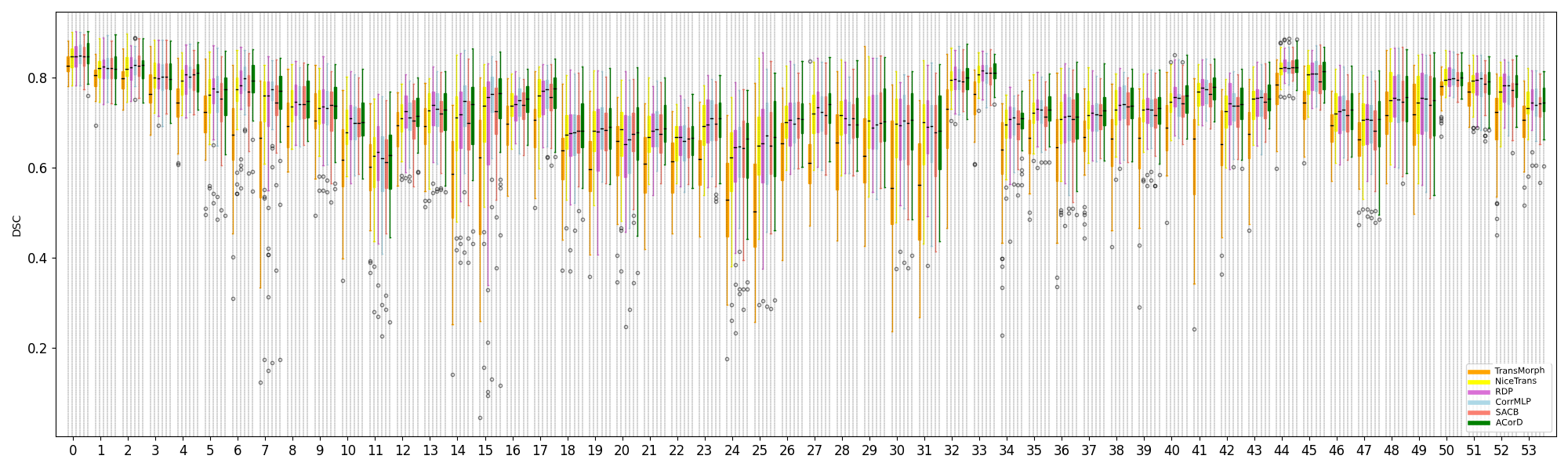}
\caption{Dice score evaluation on LPBA brain image registration (54 ROIs).}
\label{fig:box_plot_lpba}
\end{figure*}

Fig. \ref{fig:box_plot_lpba} presents the dice score evaluation results on different anatomical classes of LPBA dataset.

\section{Limitations of ACorD}
\begin{figure}[h]
\centerline{\includegraphics[width=\columnwidth]{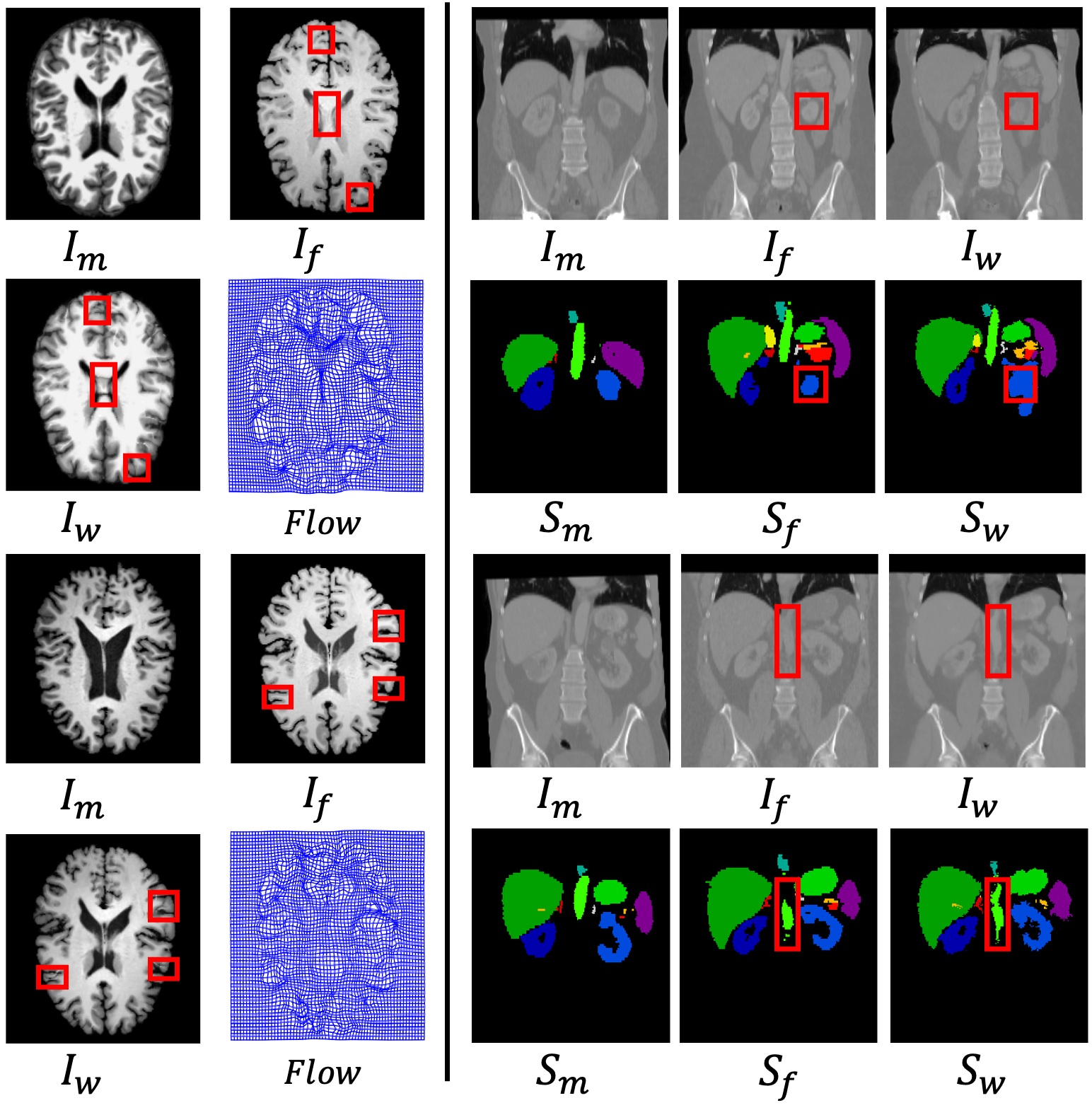}}
\caption{\textbf{Artifact of ACorD registration.} Left: In regions highlighted by the red boxes, the cortical surface is shown to have unnecessary folding. Right: While intensity similarity metrics may indicate good alignment, errors revealed through segmentation comparisons demonstrate the shortcomings of structure-unaware supervision. The mismatches show that the model might achieve high image similarity at the cost of structural integrity, leading to inaccuracies at the structure level.}
\label{fig:supp_limit}
\end{figure}

Are presented in Fig. \ref{fig:supp_limit}.

\section{Computational Cost Analysis}
Is presented in Table \ref{tab:complexity}.
\begin{table}[h]
\centering
\caption{Computational cost comparison on A6000.}
\label{tab:complexity}
{\scriptsize
\setlength{\tabcolsep}{7pt}
\renewcommand{\arraystretch}{0.7}
\begin{tabular}{l|ccc}
\toprule
\textbf{Model} & \textbf{MACs (GMac)} & \textbf{Runtime (ms)} & \textbf{Memory (MB)} \\
\midrule
TransMorph & 714.16 & 161.15 & 5353.16 \\
TransMatch & 717 & 257 & 4768 \\
TransDCA & 4766 & 593 & 9431 \\
NICE-Trans & 313 & 415 & 6205 \\
RDPNet & 3109 & 428 & 9306 \\
Dual-PR++ & 1773 & 491 & 9461 \\
Im2Grid & 73 & 367 & 10906 \\
ModeT & 83 & 870 & 11166 \\
CorrMLP & 1117 & 1237 & 7004 \\
SACB-Net & 181 & 1119 & 2836 \\
\textbf{ACorD} & \textbf{1781} & \textbf{710} & \textbf{16946} \\
\bottomrule
\end{tabular}
}
\end{table}

\begin{table}[t]
\centering
\caption{Per-scale DSC on AbdomenCT}
\label{tab:per_scale_accuracy}
{\scriptsize
\setlength{\tabcolsep}{5pt}
\renewcommand{\arraystretch}{0.65}
\begin{tabular}{l|ccccc}
\toprule
\textbf{Model}
& \textbf{L5}
& \textbf{L4}
& \textbf{L3}
& \textbf{L2}
& \textbf{L1} \\
\midrule

Exp.8 (Default+Attn)
& 0.298
& 0.403
& 0.476
& 0.534
& 0.562 \\

Exp.7 (Learned+Avg)
& 0.289
& 0.403
& 0.472
& 0.554
& 0.587 \\

ACorD
& \textbf{0.304}
& \textbf{0.408}
& \textbf{0.500}
& \textbf{0.563}
& \textbf{0.597} \\

\bottomrule
\end{tabular}
}
\end{table}

\section{Per-scale Dice on AbdomenCT}
Is presented in Table \ref{tab:per_scale_accuracy}.


\end{document}